\documentclass{article}

     \PassOptionsToPackage{numbers, compress}{natbib}

     \usepackage[preprint]{iai_neurips_2024}

\usepackage[utf8]{inputenc} 
\usepackage[T1]{fontenc}    
\usepackage{hyperref}       
\usepackage{url}            
\usepackage{booktabs}       
\usepackage{amsfonts}       
\usepackage{nicefrac}       
\usepackage{microtype}      
\usepackage{xcolor}         
\usepackage{hyperref}
\hypersetup{
    colorlinks,
    linkcolor={red!50!black},
    citecolor={blue!50!black},
    urlcolor={blue!80!black}
}
\usepackage{amsmath}
\usepackage{graphicx}
\usepackage{caption}
\usepackage{subcaption}
\usepackage{comment}
\usepackage{multirow}
\usepackage{mathtools}
\hypersetup{
    colorlinks=true,
    linkcolor=cyan
    }

\usepackage{array}
\newcommand\Tstrut{\rule{0pt}{2.6ex}}         
\newcommand\Bstrut{\rule[-0.9ex]{0pt}{0pt}}   

\title{Quantifying and Enabling the Interpretability of CLIP-like Models}

\author{%
    Avinash Madasu \\ Intel Labs \\ \texttt{avinash.madasu@intel.com} \And Yossi Gandelsman \\ UC Berkeley \\ \texttt{yossi@gandelsman.com} \AND  Vasudev Lal \\ Intel Labs \\ \texttt{vasudev.lal@intel.com} \And Phillip Howard  \\ Intel Labs \\ \texttt{phillip.r.howard@intel.com} \\
}

\begin{document}

\maketitle

\begin{abstract}
  CLIP is one of the most popular foundational models and is heavily used for many vision-language tasks. However, little is known about the inner workings of CLIP.  To bridge this gap we propose a study to quantify the interpretability in CLIP like models. We conduct this study on six different CLIP models from OpenAI and OpenCLIP which vary by size, type of pre-training data and patch size. Our approach begins with using the TEXTSPAN algorithm and in-context learning to break down individual attention heads into specific properties. We then evaluate how easily these heads can be interpreted using new metrics which measure property consistency within heads and property disentanglement across heads. Our findings reveal that larger CLIP models are generally more interpretable than their smaller counterparts. To further assist users in understanding the inner workings of CLIP models, we introduce CLIP-InterpreT, a tool designed for interpretability analysis. CLIP-InterpreT offers five types of analyses: property-based nearest neighbor search, per-head topic segmentation, contrastive segmentation, per-head nearest neighbors of an image, and per-head nearest neighbors of text.
\end{abstract}

\section{Introduction}
CLIP ~\cite{radford2021learning}, a large-scale vision-language (VL) model, is extensively used as a foundational model for tasks such as video retrieval, image generation, and segmentation ~\cite{luo2022clip4clip, liu2024improved, brooks2023instructpix2pix, esser2024scaling, kirillov2023segment}. Given its widespread use, it is imperative to understand the inner workings of CLIP. Towards this end, we introduce a systematic methodology for quantifying the interpretability in CLIP like models. We perform this study on six types of CLIP models: ViT-B-16, ViT-B-32, and ViT-L-14 from OpenAI, as well as ViT-B-16, ViT-B-32, and ViT-L-14 from OpenCLIP \cite{ilharco_gabriel_2021_5143773}. 

First, we identify interpretable structures within the individual heads of the last four layers of the model, using a set of text descriptions. To accomplish this, we employ the \textsc{TextSpan} algorithm~\cite{gandelsmaninterpreting}, which helps us find the most appropriate text descriptions for each head. 
After identifying these text descriptions, we assign labels to each head, representing the common property shared by the descriptions. This labeling process is carried out using in-context learning with ChatGPT. We begin by manually labeling five pairs of text descriptions and their corresponding property labels, which serve as examples. These examples are then used to prompt ChatGPT to assign labels for the remaining heads. This approach systematically connects the attention heads to the properties they learn during large-scale pre-training, offering insights into the roles of individual heads.

We introduce two metrics, the entanglement score and the association score, to quantify interpretability in CLIP models. These metrics are specifically designed to assess how easily properties can be linked to each attention head within the model. The entanglement score highlights that in larger CLIP models, attention heads tend to exhibit greater independence, meaning they seldom share properties with other heads. This indicates a clearer distinction in the roles of individual heads. Similarly, the association score shows that larger CLIP models consistently focus on a single property within the associated text descriptions, reinforcing the idea that these models are more interpretable because their heads are more specialized and less likely to overlap in the properties they learn. 

Finally, we propose a new interpretability user application: CLIP-InterpreT (Figure~\ref{fig:clipinterpret-interface}) designed specifically to help users understand the inner workings of CLIP. This tool is engineered to adapt and integrate five distinct interpretability analysis methods, allowing for comprehensive insights into six different variants of CLIP models. 

\section{Related work}
Early research on interpretability primarily concentrated on convolutional neural networks (CNNs) due to their intricate and opaque decision-making processes \citep{zeiler2014visualizing,selvaraju2017grad,simonyan2014deep,fong2017interpretable,hendricks2016generating}. 
More recently, the interpretability of Vision Transformers (ViT) has garnered significant attention as these models, unlike CNNs, rely on self-attention mechanisms rather than convolutions. Researchers have focused on task-specific analyses in areas such as image classification, captioning, and object detection to understand how ViTs process and interpret visual information \cite{dong2022towards, elguendouze2023explainability, mannix2024scalable, xue2022protopformer, cornia2022explaining, dravid2023rosetta}. One of the key metrics used to measure interpretability in ViTs is the attention mechanism itself, which provides insights into how the model distributes focus across different parts of an image when making decisions \cite{cordonnier2019relationship, chefer2021transformer}. This has led to the development of techniques that leverage attention maps to explain ViT predictions. Early work on multimodal interpretability, which involves models that handle both visual and textual inputs, probed tasks such as how different modalities influence model performance \cite{cao2020behind, madasu2023multimodal} and how visual semantics are represented within the model \cite{hendricks2021probing, lindstrom2021probing}. Aflalo et al. \citep{aflalo2022vl} explored interpretability methods for vision-language transformers, examining how these models combine visual and textual information to make joint decisions. Similarly, Stan et al. \citep{stan2024lvlm} proposed new approaches for interpreting vision-language models, focusing on the interactions between modalities and how these influence model predictions. Our work builds upon and leverages the methods introduced by \citet{gandelsmaninterpreting,gandelsman2024neurons} to interpret attention heads, neurons, and layers in vision-language models, providing deeper insights into their decision-making processes. 


\section{Methodology}
In this section, we outline the methodology used in our analysis, beginning with an explanation of the \textsc{TextSpan} algorithm. We then describe how we extend this algorithm to apply it across all attention heads in multiple CLIP models using in-context learning.

The \textsc{TextSpan} algorithm ~\cite{gandelsmaninterpreting} is designed to decompose individual attention heads by associating them with corresponding text descriptions. It requires an initial set of text descriptions broadly capturing the concepts in an image. The algorithm starts by generating two matrices: $C$, which contains the outputs for a specific head (denoted as $(l, h)$), and $R$, which includes representations of candidate text descriptions projected onto the span of $C$.
In each iteration, \textsc{TextSpan} calculates the dot product between each row of $R$ and the outputs in $C$ to identify the row with the highest variance, known as the "first principal component." Once identified, this component is projected away from all rows, and the process is repeated to find additional components. The projection step ensures that each new component adds variance that is orthogonal to the earlier components, thereby isolating distinct aspects of the text descriptions relevant to each head.

\textsc{TextSpan} is effective at identifying text descriptions that are most similar to a given head. To label the common property shared by these text descriptions, we employ in-context learning with ChatGPT. We begin by manually labeling properties for five heads, which serve as examples. These examples are then used to prompt ChatGPT to generate labels for the remaining heads. This approach allows us to systematically label the properties associated with each head.

\section{Quantifying interpretability in CLIP models}
Given the numerous CLIP-like models which are available, a key question arises: can we quantify how interpretable these models are? 
To answer this, we introduce a set of metrics specifically designed to assess how easily properties can be linked to each layer and head within the models. These metrics provide a way to measure the interpretability of the models, helping us understand how clearly different properties are represented.
\subsection{Entanglement score}
We define the entanglement score as the mean frequency at which heads exhibit the same \textsc{TextSpan} label (property). A higher score suggests that the property is shared among multiple heads or layers, making it more challenging to distinctly associate the property with a particular head. If the score is zero, then all the heads are disentangled from each other. 

Table~\ref{tab: entanglement-score} presents the entanglement scores across various CLIP models, revealing several key insights. Notably, larger CLIP models exhibit significantly lower entanglement scores compared to their base model counterparts, suggesting a more distinct association of properties to specific heads and layers in these larger models. This reduced entanglement is an important result as it highlights the improved interpretability and clarity in larger models.

Additionally, the table highlights another key observation: OpenAI’s smaller models are less entangled than the models from OpenCLIP, indicating a potential difference in how these models manage property associations at a smaller scale. Conversely, the entanglement scores for OpenAI’s larger models are higher than expected, showing more entanglement than their smaller versions, which could suggest a trade-off in complexity and property association within these models. These findings emphasize the importance of model size and architecture in determining the level of entanglement,
\begin{table*}
	\begin{center}
		\resizebox{0.75\textwidth}{!}
		{
		\begin{tabular}{c c c c c} %
	    \hline
         \textbf{Model\Tstrut\Bstrut} & \textbf{Model size\Tstrut\Bstrut} & \textbf{Patch size\Tstrut\Bstrut} & \textbf{Pre-training data\Tstrut\Bstrut} & \textbf{Entanglement Score ($\downarrow$) \Tstrut\Bstrut} \\
        \hline
CLIP\Tstrut & B \Tstrut  & 32\Tstrut & OpenAI-400M\Tstrut & 0.437\Tstrut   \\
CLIP\Tstrut & B \Tstrut  & 32\Tstrut & OpenCLIP-datacomp\Tstrut & 0.52\Tstrut   \\
CLIP\Tstrut & B \Tstrut  & 16\Tstrut & OpenAI-400M\Tstrut & 0.5\Tstrut   \\
CLIP\Tstrut & B \Tstrut  & 16\Tstrut & OpenCLIP-LAION2B\Tstrut & 0.541\Tstrut   \\
CLIP\Tstrut & L \Tstrut  & 14\Tstrut & OpenAI-400M\Tstrut & 0.359\Tstrut   \\
CLIP\Tstrut & L \Tstrut  & 14\Tstrut & OpenCLIP-LAION2B\Tstrut & 0.343\Tstrut   \\
 \hline
		\end{tabular}}
		\caption{\textbf{Entanglement scores for CLIP models}. Larger CLIP models are less entangled.}
        \label{tab: entanglement-score}
	\end{center}
\end{table*}

\subsection{Association score}
Previously, we explored how the \textsc{TextSpan} labels assigned to different attention heads in a model can reflect the model's interpretability. This led to the question of whether all the text descriptions linked to a given head genuinely correspond to the assigned property label. To investigate this, we manually evaluated how frequently the text descriptions align with the property label. We introduced the "association score" to quantify this, which is defined as the average frequency of heads with at least three text descriptions matching the property label.
Table~\ref{tab: association-score} presents the association scores for various CLIP models. The data clearly shows that larger CLIP models have more heads consistently focusing on a single property, making them more interpretable. This observation is consistent with the results from the entanglement scores, which lead to a similar conclusion. 

From Tables~\ref{tab: entanglement-score} and~\ref{tab: association-score}, it is evident that larger models have heads that learn properties independently of other heads while consistently focusing on a single property. This independence allows for the easy isolation of head-property pairs, enhancing the interpretability of the model's individual heads.
Additionally, OpenCLIP’s smaller models have lower association scores compared to their OpenAI counterparts, while the opposite is true for larger models. This pattern mirrors the findings seen in the entanglement scores, reinforcing the insights gained from both metrics.

\begin{table*}
	\begin{center}
		\resizebox{0.75\textwidth}{!}
		{
		\begin{tabular}{c c c c c} %
	    \hline
         \textbf{Model\Tstrut\Bstrut} & \textbf{Model size\Tstrut\Bstrut} & \textbf{Patch size\Tstrut\Bstrut} & \textbf{Pre-training data\Tstrut\Bstrut} & \textbf{Association Score ($\uparrow$) \Tstrut\Bstrut} \\
        \hline
CLIP\Tstrut & B \Tstrut  & 32\Tstrut & OpenAI-400M\Tstrut & 0.437\Tstrut   \\
CLIP\Tstrut & B \Tstrut  & 32\Tstrut & OpenCLIP-datacomp\Tstrut & 0.145\Tstrut   \\
CLIP\Tstrut & B \Tstrut  & 16\Tstrut & OpenAI-400M\Tstrut & 0.354\Tstrut   \\
CLIP\Tstrut & B \Tstrut  & 16\Tstrut & OpenCLIP-LAION2B\Tstrut & 0.166\Tstrut   \\
CLIP\Tstrut & L \Tstrut  & 14\Tstrut & OpenAI-400M\Tstrut & 0.453\Tstrut   \\
CLIP\Tstrut & L \Tstrut  & 14\Tstrut & OpenCLIP-LAION2B\Tstrut & 0.562\Tstrut   \\
 \hline
		\end{tabular}}
		\caption{\textbf{Association scores for CLIP models}. Larger models have greater property consistency.}
        \label{tab: association-score}
	\end{center}
\end{table*}

\section{CLIP-InterpreT}
\begin{figure}
    \centering
    \includegraphics[width=0.9\linewidth]{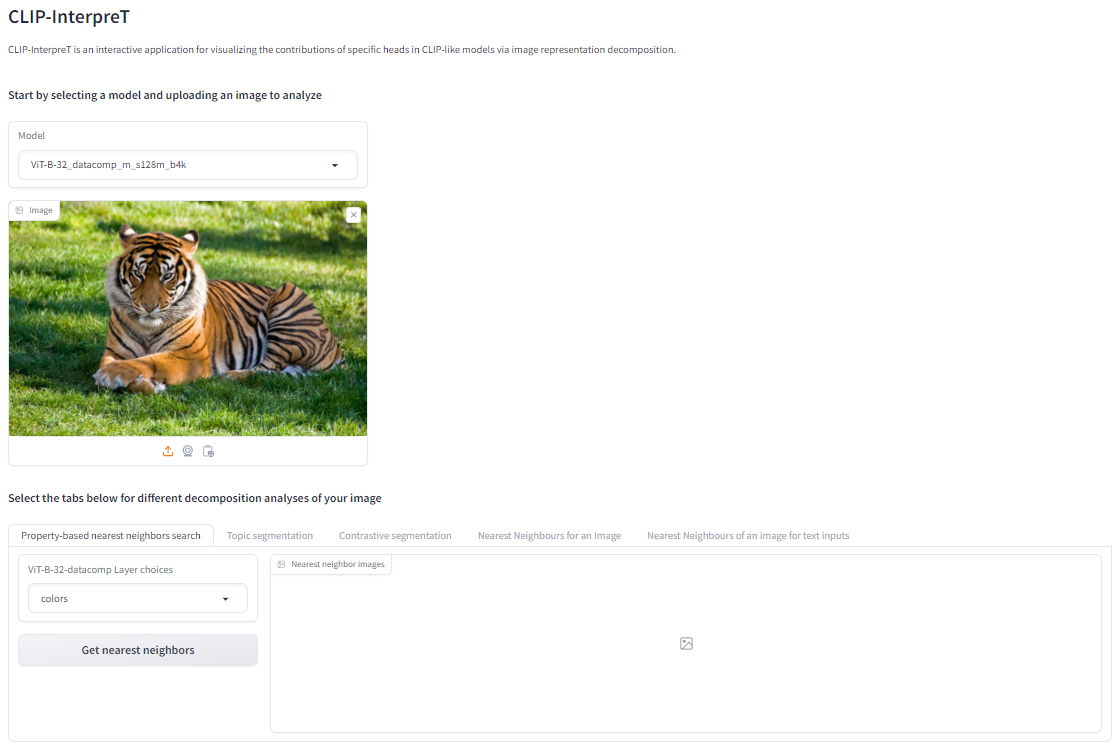}
    \caption{Interface of CLIP-InterpreT application. Users can upload an image and select a model to analyze. There are five tabs for different decomposition analyses.}
    \label{fig:clipinterpret-interface}
\end{figure}

In the previous sections, we outlined a systematic approach to quantifying interpretability in CLIP models. However, the true value of interpretability lies in its accessibility to users. To address this, we introduce a new application called CLIP-InterpreT, a comprehensive tool designed to empower users with insights into the inner workings of CLIP-like models. This application provides an intuitive interface, as shown in Figure~\ref{fig:clipinterpret-interface}. In the user interface, users can easily upload an image of their choice and select one of six CLIP models for analysis. These models include ViT-B-16, ViT-B-32, and ViT-L-14 from OpenAI, as well as ViT-B-16, ViT-B-32, and ViT-L-14 from OpenCLIP \cite{ilharco_gabriel_2021_5143773}. The application offers five distinct types of interpretability analyses, which we will discuss next.

\begin{figure}
    \centering
    \includegraphics[width=0.8\linewidth]{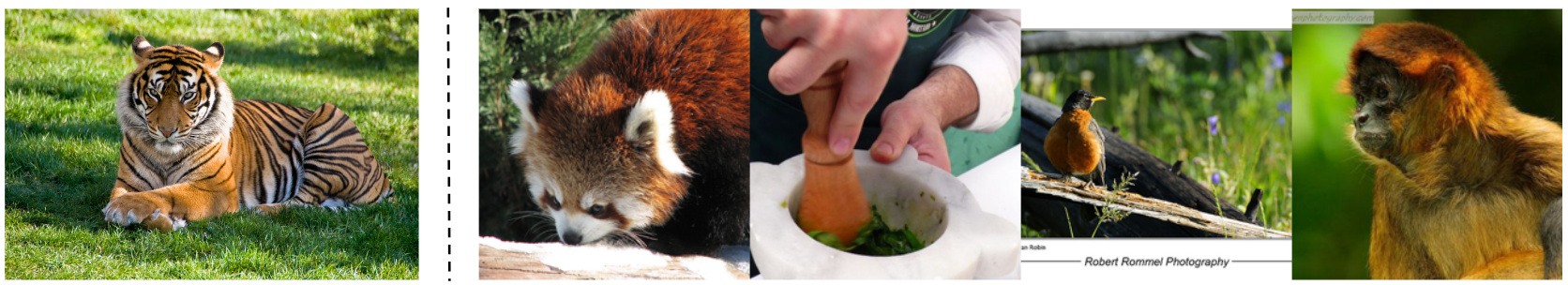}
    \caption{\textbf{Top-4 nearest neighbors for "colors" property}. The model used is ViT-B-32 (Data comp). The input is the image of tiger which is on the left of the dotted lines. The outputs are the four images, right of the dotted lines. In this example, we see that both the input and retrieved output images have common orange, black, and green colors.}
    \label{fig:property-nearest-tiger-colors}
\end{figure}
\subsection{Property-based nearest neighbors search}
In this analysis, we demonstrate that the layers and heads of CLIP models can be characterized by specific attributes such as colors, locations, animals, and more. Since multiple heads can learn the same attribute, we combine the representations of these heads, which have been labeled using in-context learning, to create unified image representations. To evaluate the similarity between an input image and a set of candidate images from the ImageNet validation dataset \cite{deng2009imagenet}, we calculate the cosine similarity between the unified representation of the input image and each candidate image. The top four images with the highest similarity scores are then retrieved, showing which images are most similar to the input image based on the learned attributes.
Figure~\ref{fig:property-nearest-tiger-colors} provides an example focusing on the color property which was obtained from CLIP-InterpreT. The model used in this example is ViT-B-32 pretrained on datacomp \cite{ilharco_gabriel_2021_5143773}. The input image uploaded to CLIP-InterpreT, located to the left of the dotted line, features a tiger. On the right side of the dotted line are the four images retrieved by the model. In this case, we observe that both the input image and the retrieved images share common colors, specifically orange, black, and green.

\begin{figure}
    \centering
    \includegraphics[width=0.8\linewidth]{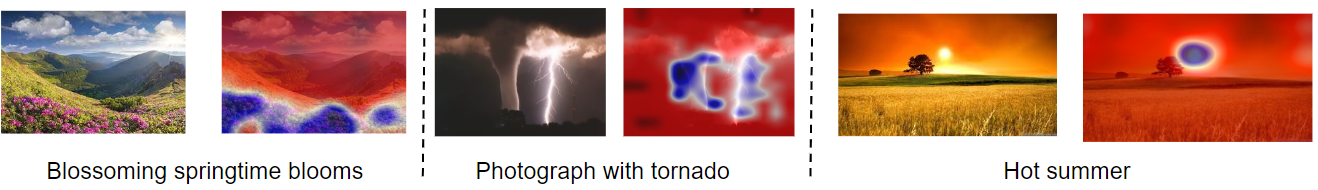}
    \caption{\textbf{Topic Segmentation results for Layer 11, Head 3 (an "environment/weather" head).}. The model used is ViT-B-16 (LAION-2B). In the first image (left), the heatmap (blue) is focused on "flowers" which matches the text description. In the second image (middle), the heatmap (blue) is concentrated on the "tornado" matching the text description. In the last image, the heatmap (blue) is focused on "sun" matching the description "Hot Summer".}
    
    \label{fig:vit-b-16-openai-environment-l11-h3-topic-seg}
\end{figure}
\subsection{Per head topic Segmentation}

In contrast to \citet{gandelsmaninterpreting}, we study topic segmentation for each individual head in various CLIP-like models using CLIP-InterpreT. This analysis focuses on projecting a segmentation map corresponding to an input text onto the reference image. The segmentation map is computed using various heads of the model to highlight the specific properties that each head captures, which allows us to visualize how different heads focus on different elements within the image based on the given text description.
Figure~\ref{fig:vit-b-16-openai-environment-l11-h3-topic-seg} provides examples obtained using CLIP-InterpreT for layer 11, head 3 of the ViT-B-16 model, which has been pretrained on the LAION-2B dataset. In the first image on the left, the heatmap (represented in blue) focuses on the "flowers," aligning well with the text description. In the second image, positioned in the middle, the heatmap (blue) concentrates on the "tornado," again corresponding to the text. Finally, in the third image, the heatmap (blue) is centered on the "sun," accurately reflecting the description "Hot Summer." These examples illustrate how the model's attention is directed towards specific features in the image that match the input text.

\begin{figure}
    \centering
    \includegraphics[width=0.8\linewidth]{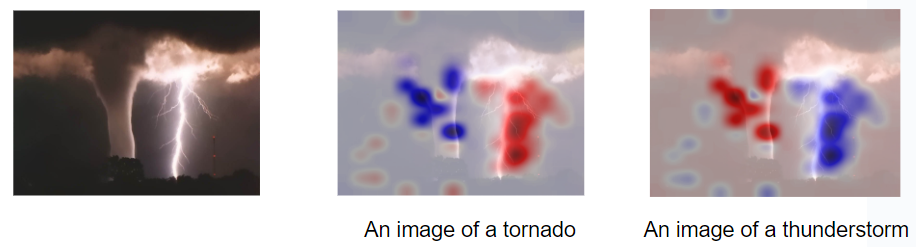}
    \caption{\textbf{Image shows the contrastive Segmentation between portions of the image containing "tornado" and "thunderstorm."} The model used is ViT-L-14 pretrained on LAION-2B dataset.}
    \label{fig:thunder_contrastive_seg}
\end{figure}
\subsection{Contrastive Segmentation}
The contrastive segmentation analysis provided in CLIP-InterpreT contrasts the visual interpretations of a single image when described by two different text inputs. By projecting the segmentation maps corresponding to each input onto the original image, we reveal how the model visually interprets and differentiates the input texts. Figure~\ref{fig:thunder_contrastive_seg} provides an example using the ViT-L-14 model, pretrained on the LAION-2B dataset. The image illustrates the contrast between segments corresponding to "tornado" and "thunderstorm", in the image showcasing how the model distinguishes between these two concepts.

\begin{figure}
    \centering
    \includegraphics[trim={0 0 0 6.7cm},clip,width=0.8\linewidth]{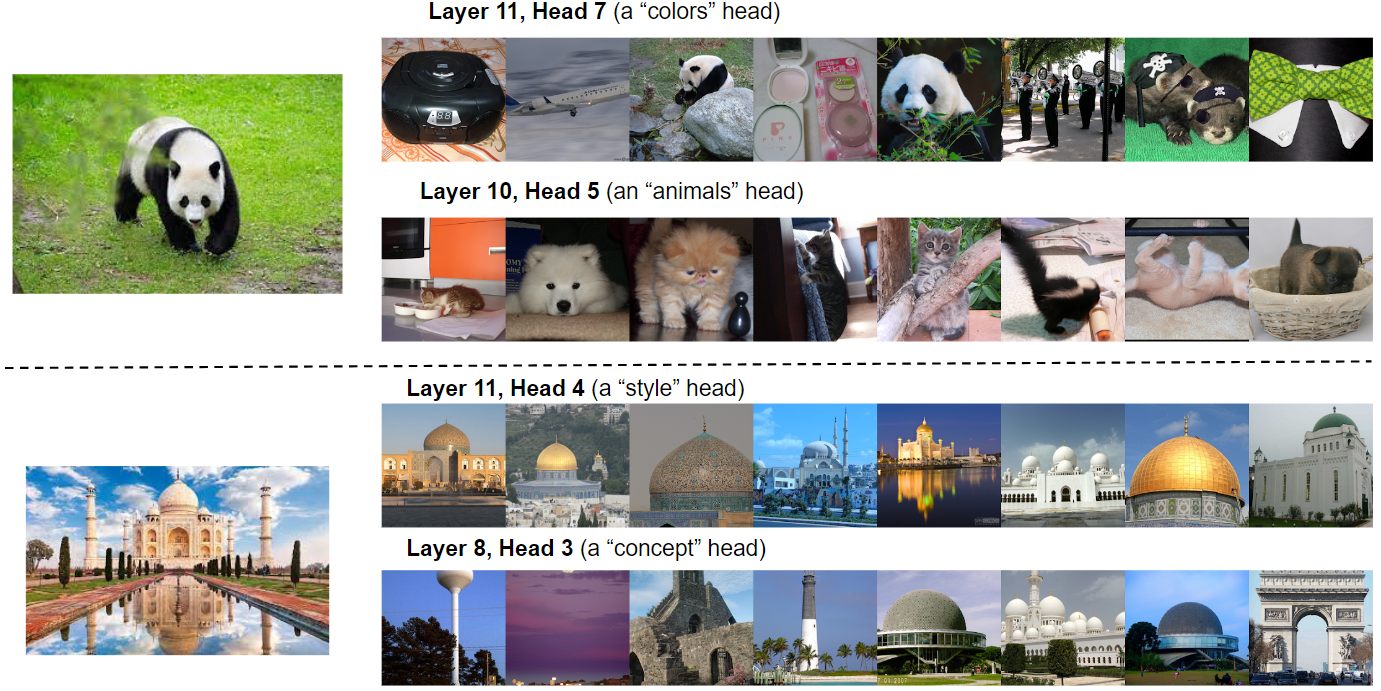}
    \caption{\textbf{Top-8 nearest neighbors per head and image}. The input image is provided on the left, with the head-specific nearest neighbors shown on the right. The model used is OpenAI-400M.}
    \label{fig:vit-b-16-openai-nearest-image}
\end{figure}
\subsection{Per-head Nearest Neighbors of an Image}
CLIP-InterpreT can also visualize the nearest neighbors retrieved for an input image based on similarity scores computed using a single attention head. Certain heads are specialized in capturing specific image properties, allowing us to leverage their intermediate representations for better interpretability. By calculating the similarity of direct contributions from individual heads, we can identify images that closely match the input image in terms of specific aspects, such as colors or objects.
Figure~\ref{fig:vit-b-16-openai-nearest-image} illustrates this concept using the ViT-B-16 model from OpenAI. For an image of pandas which was uploaded to CLIP-InterpreT, the eight nearest neighbors retrieved from layer 11, head 7 (which focuses on 'colors'), display similar color patterns. Meanwhile, layer 10, head 5 (which specializes in 'animals'), retrieves images featuring animals. Similarly, for an input image of the Taj Mahal, layer 11, head 4 (the 'style' head) retrieves images with a similar architectural style, while layer 8, head 3 returns images that share a common concept, such as structures set against open skies.

\begin{figure}
    \centering
    \includegraphics[width=0.8\linewidth]{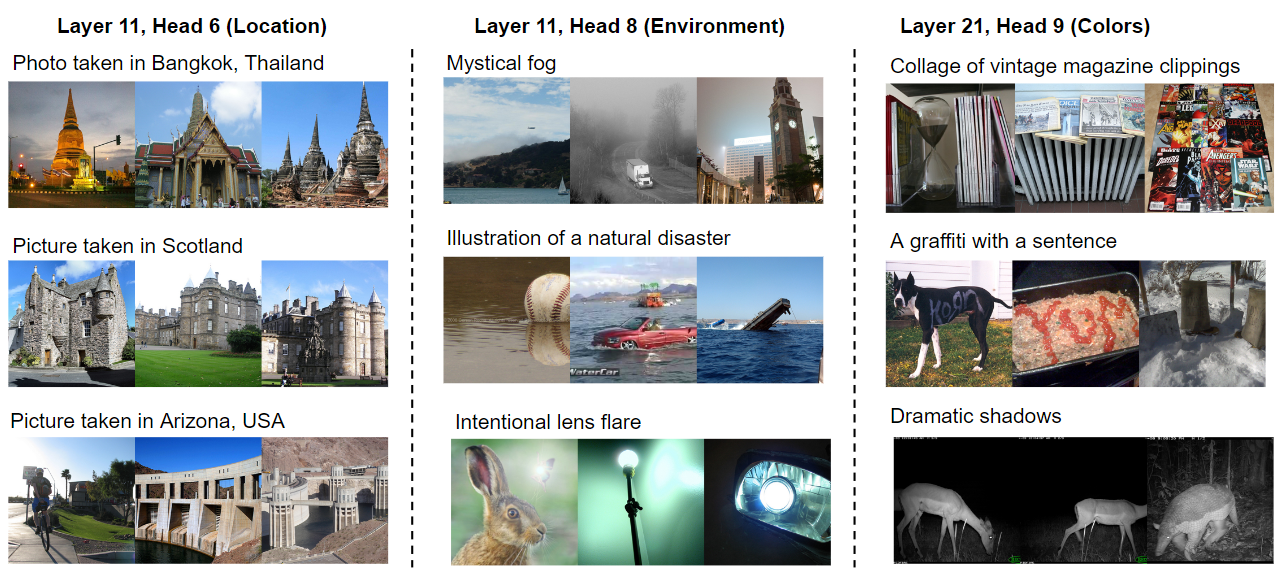}
    \caption{\textbf{Nearest neighbors retrieved for the top \textsc{TextSpan} outputs of a given layer and head}. The model used is ViT-B-16 pretrained on OpenAI-400M.}
    \label{fig:Vit-b-16-openai-nearest-text}
\end{figure}
\subsection{Per-head Nearest neighbors of a Text Input}
To determine whether CLIP-like models can link image representations to a given text input, CLIP-InterpreT provides the ability to retrieve the nearest neighbors for a given input text using different attention heads. We utilize the top \textsc{TextSpan} outputs identified for each head. Figure~\ref{fig:Vit-b-16-openai-nearest-text} provides an example of this analysis. In the figure, we observe that layer 11, head 6 (the 'location' head) retrieves images that match the text caption. We also see similar behavior from layer 11, head 8 (the 'environment' head), and layer 21, head 8 (the 'colors' head).

\section{Conclusion}
In this work, we conducted an analysis to quantify the interpretability of six different CLIP-like models from OpenAI and OpenCLIP, varying in size, pre-training data, and patch size. We used the \textsc{TextSpan} algorithm and in-context learning to decompose individual attention heads into specific properties and then evaluated interpretability using newly developed metrics. Our findings show that larger CLIP models are generally easier to interpret than smaller ones. To help users explore these insights, we introduced CLIP-InterpreT, a tool offering five types of interpretability analyses.
\bibliographystyle{plainnat}
\bibliography{egbib}


\newpage
\appendix

\section{Appendix / supplemental material}
In this section, we present additional figures from our CLIP-InterpreT application.

\begin{figure}[ht]
    \centering
    \includegraphics[width=\linewidth]{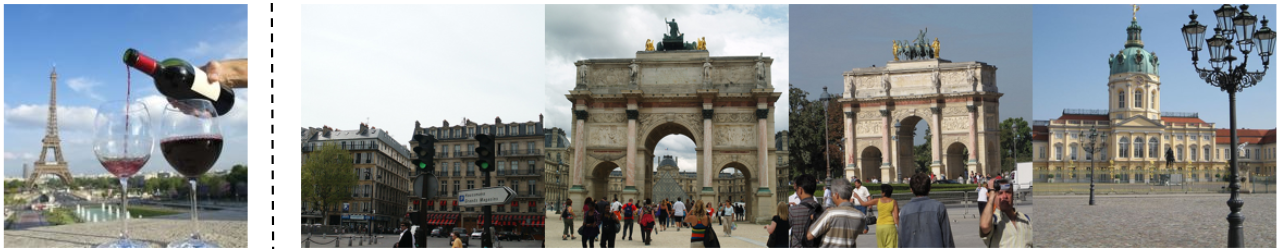}
    \caption{\textbf{Top-4 nearest neighbors for "location" property.} The model used is ViT-B-32 (OpenAI). The input image is a picture of the "Eiffel tower" in Paris, France. The top-4 images are related to popular location landmarks.}
    \label{fig:France_location_vit-b-32-openai_combined}
\end{figure}

\begin{figure}[ht]
    \centering
    \includegraphics[width=\linewidth]{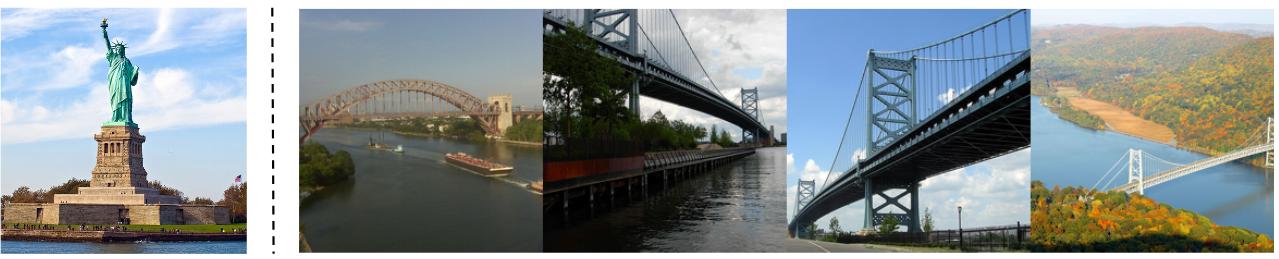}
    \caption{\textbf{Top-4 nearest neighbors for "location" property.} Top-4 nearest neighbors for "location" property. The model used is ViT-L-14 (LAION2B).}
    \label{fig:Statue_of_liberty_location_ViT-L-14-laion_combined}
\end{figure}

\begin{figure}[ht]
    \centering
    \includegraphics[width=\linewidth]{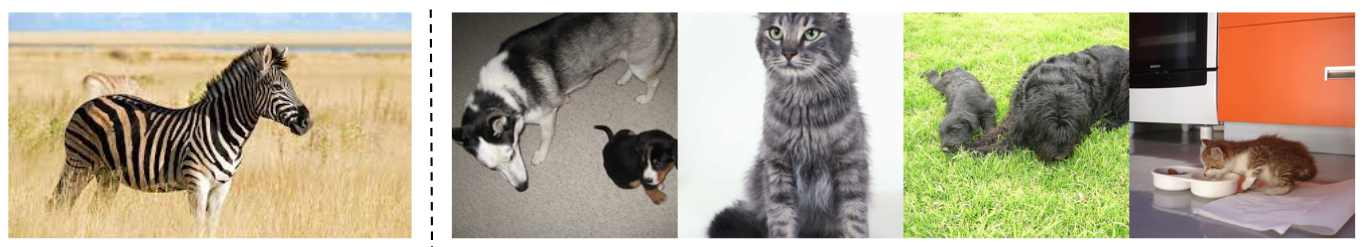}
    \caption{\textbf{Top-4 nearest neighbors for "animals" property.} The model used is ViT-B-16 (OpenAI).}
    \label{fig:Animals_zebra_vit-b-16-openai-combined}
\end{figure}

\begin{figure}[htbp]
    \centering
    \includegraphics[width=\linewidth]{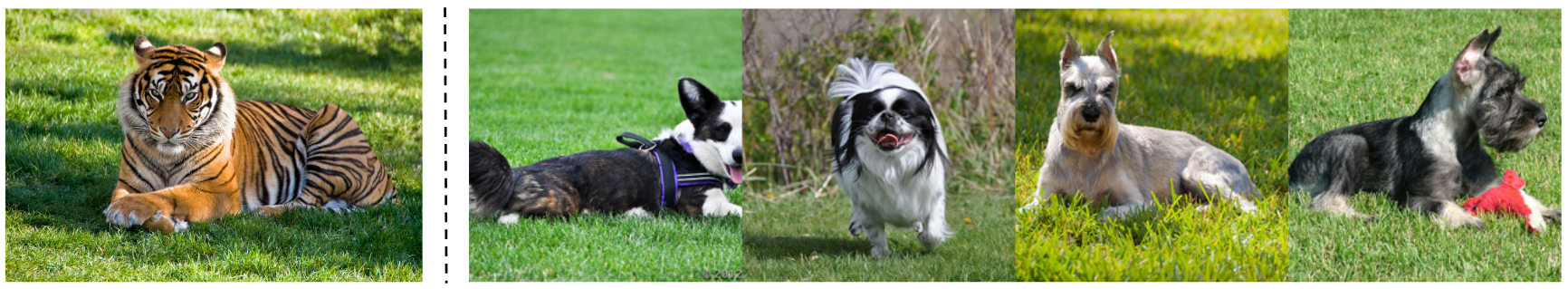}
    \caption{\textbf{Top-4 nearest neighbors for "pattern" property.} The model used is ViT-B-32 (OpenAI). In this example, the input image shows an animal laying on the grass. The top retrieved images also show this common pattern of an animal laying on the grass.}
    \label{fig:Animals_zebra_vit-b-16-openai-combined}
\end{figure}

\begin{figure}
    \centering
    \includegraphics[width=\linewidth]{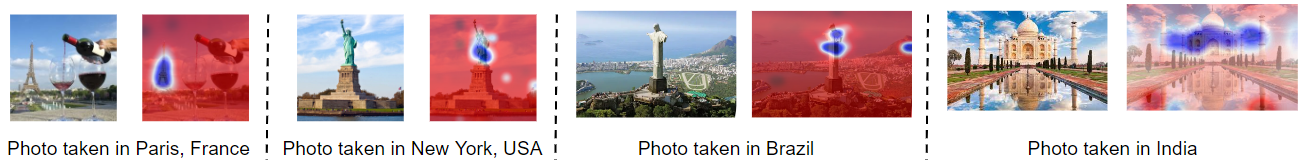}
    \caption{\textbf{Topic Segmentation results for Layer 22, Head 13 (a "geolocation" head).} The model utilized is ViT-L-14 (trained on LAION-2B). The blue highlight in the segmentation map focuses on landmarks such as the "Eiffel Tower," "Christ the Redeemer," "Statue of Liberty," and "Taj Mahal," which are located in France, New York, Brazil, and India, respectively, as indicated in the text input. Notably, the text does not explicitly state that the Eiffel Tower is in Paris, France. Instead, Layer 22, Head 13 of the model possesses geolocation properties that implicitly identify these locations.}
    \label{fig:Vit-l-14-laoin-location-l22-h13-topic-seg}
\end{figure}

\begin{figure}
    \centering
    \includegraphics[width=\linewidth]{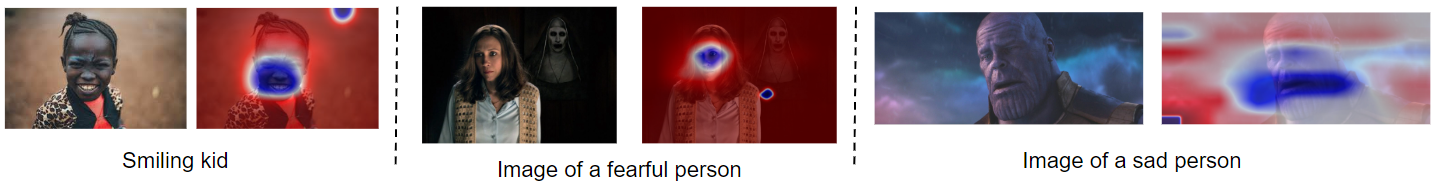}
    \caption{\textbf{Topic Segmentation results for Layer 10, Head 6 (an "emotion" head).} The model employed is ViT-B-32 (OpenAI-400M). In the first image (on the left), the heatmap (in blue) prominently highlights the "smile" emotion on a child's face, aligning well with the text description. The middle image shows the heatmap concentrated on the "fear" emotion, which is associated with the Conjuring movie. Notably, there is no explicit indication in the text that the image is meant to convey "fear." In the final image, the heatmap centers on the "sad" emotion displayed by "Thanos" from Marvel.}
    \label{fig:Vit-l-14-laoin-location-l22-h13-topic-seg}
\end{figure}

\begin{figure}
    \centering
    \includegraphics[width=\linewidth]{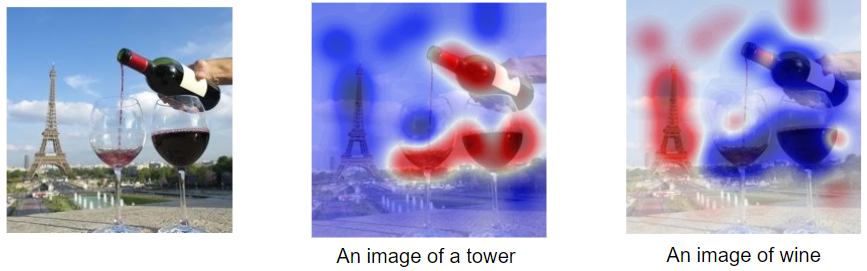}
    \caption{The image illustrates a contrastive segmentation between areas of the image associated with "tower" and "wine." The model utilized for this analysis is ViT-L-16 (trained on LAION-2B).}
    \label{fig:Vit-b-16-laoin-france-contrastive-seg}
\end{figure}

\begin{figure}
    \centering
    \includegraphics[width=\linewidth]{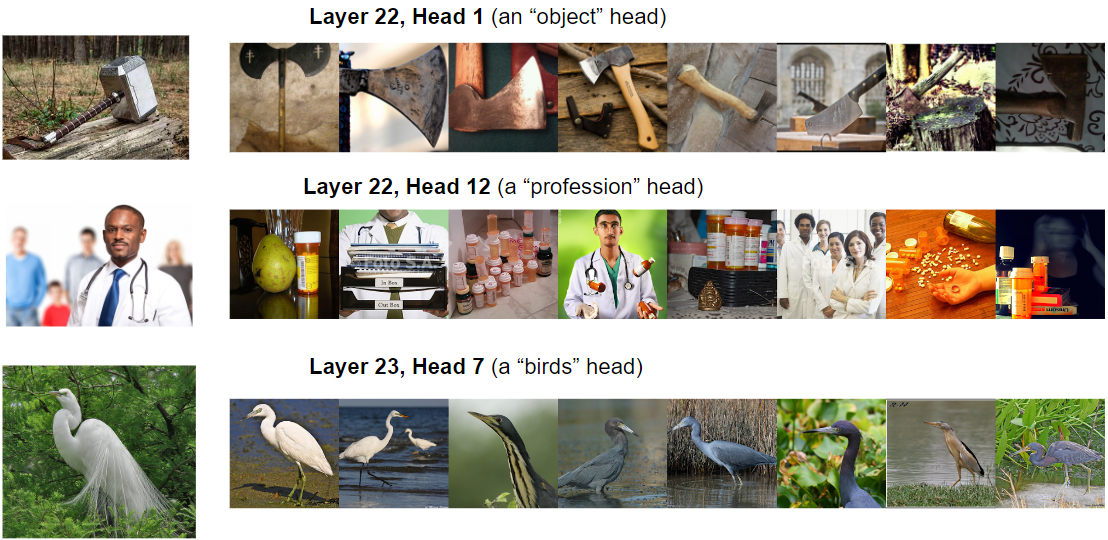}
    \caption{\textbf{Top-8 nearest neighbors per head and image.} The model used is ViT-L-14 pretrained on OpenAI-400M.}
    \label{fig:vit-l-14-openai-nearest-image}
\end{figure}

\begin{figure}
    \centering
    \includegraphics[width=\linewidth]{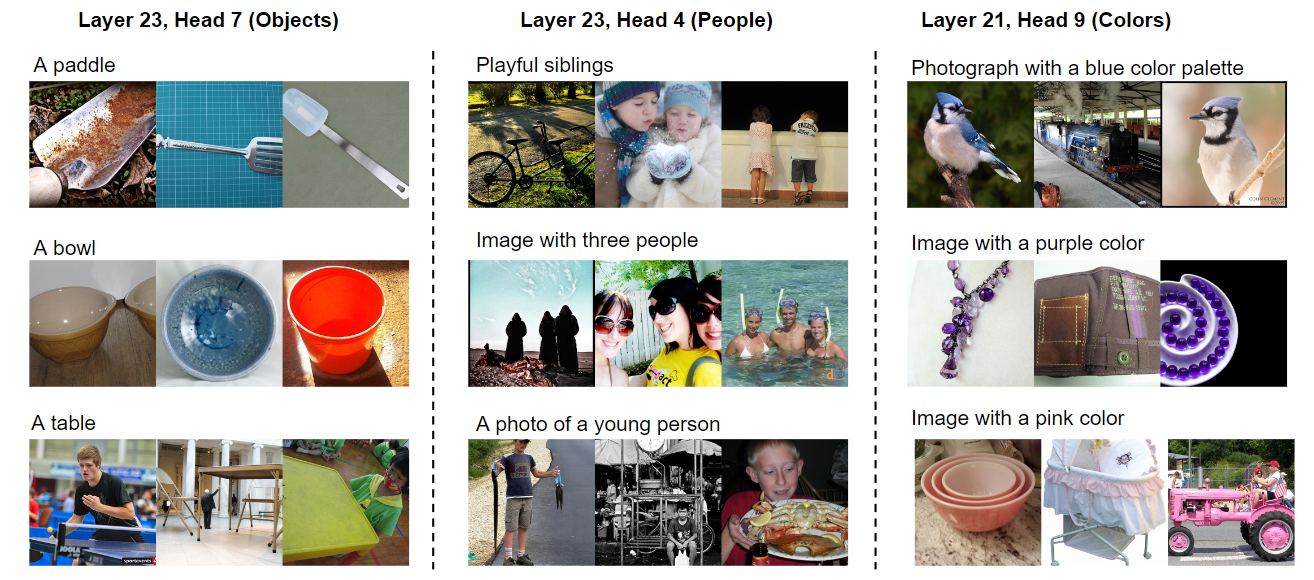}
    \caption{Nearest neighbors retrieved for the top \textsc{TextSpan} outputs of a given layer and head. The model used is ViT-L-14 pretrained on LAION-2B.}
    \label{fig:Vit-l-14-laoin-nearest-neighbours-text}
\end{figure}

\end{document}